\documentclass[11pt]{article}

\usepackage{acl}

\usepackage{times}
\usepackage{latexsym}
\usepackage{hyperref}
\usepackage{graphicx}
\usepackage{multirow}
\usepackage{booktabs}
\usepackage{bbm}
\usepackage{amsmath}
\usepackage{amssymb}
\usepackage{caption}
\usepackage{subcaption}

\usepackage[T1]{fontenc}
\usepackage[utf8]{inputenc}
\usepackage{microtype}

\title{We Need to Talk About Classification Evaluation Metrics in NLP}

\author{Peter Vickers \\
  University of Sheffield\\
  \texttt{pgjvickers1@sheffield.ac.uk} \\\And
  Loïc Barrault   \\
  Meta AI France \\
  \texttt{ loicbarrault@meta.com} \\\AND
  Emilio Monti  \\
  Amazon U.K.\\
  \texttt{monti@amazon.co.uk} \\\And
  Nikolaos Aletras   \\
  University of Sheffield\\
  \texttt{n.aletras@sheffield.ac.uk}}

\date{}

\begin{document}
\maketitle

\begin{abstract}
In Natural Language Processing (NLP) classification tasks such as topic categorisation and sentiment analysis, model generalizability is generally measured with standard metrics such as Accuracy, F-Measure, or AUC-ROC. The diversity of metrics, and the arbitrariness of their application suggest that there is no agreement within NLP on a single best metric to use. This lack suggests there has not been sufficient examination of the underlying heuristics which each metric encodes. To address this we compare several standard classification metrics with more `exotic' metrics and demonstrate that a random-guess normalised Informedness metric is a parsimonious baseline for task performance. To show how important the choice of metric is, we perform extensive experiments on a wide range of NLP tasks including a synthetic scenario,  natural language understanding, question answering and machine translation. Across these tasks we use a superset of metrics to rank models and find that Informedness best captures the ideal model characteristics.  Finally, we release a Python implementation of Informedness following the SciKitLearn classifier format.\footnote{Code and documentation is available at \url{https://github.com/petervickers/aacl-informedness}}  
\end{abstract}

\section{Introduction}
Some of the most widely used classification metrics for measuring classifier performance in NLP tasks are {\it Accuracy}, {\it F1-Measure} and the {\it Area Under the Curve - Receiver Operating Characteristics (AUC-ROC)}.  For example, seven out of nine tasks of popular NLP benchmark GLUE~\citep{wang-etal-2018-glue} use either Accuracy or F1. 

Such metrics reduce the full collection of true classes $y$ and predicted classes $\hat{y}$ to a single scalar value. For instance accuracy, the most common classification metric, is equal to the proportion of predicted classes which match true classes. Whilst capturing all the qualities of a classifier in any single scalar value is rather impossible \citep{Chicco2021}, the quality of the heuristic rule \citep{Albacete2013} influences both the overall ranking of models and the intra-task understanding of model capability.

It is difficult to evaluate true model ability with Accuracy due to the `Accuracy Paradox' \citep{David2007}: simply guessing the most common class can reward a score equal to that class's prevalence in the test set. We expand this paradox into two phenomena: (1) the reward given to models that predict more classes which appear more often (are more prevalent) \citep{Lafferty2001ConditionalRF}; and (2) the probabilistic lower bound  for accuracy being much greater than zero for random guessing models in most realistic scenarios, a phenomenon we term \emph{baseline credit} \citep{Youden1950}. 

F1-Measure \citep{Manning1999} is the harmonic mean of precision and recall and so represents a balance of two desirable characteristics of classifiers. F1 is defined against a single class, and so within even a binary classification case its value changes if the classes are reversed. Additionally, the weighting of precision and recall is a function of the model itself \citep{Hand2018}, making it a poor metric for ranking models. In order to handle the multi-class case, macro- and micro- averaging strategies have been proposed. In the single-label case we consider, micro averaging is reduced to Accuracy, whilst macro-averaging is equivalent to averaging the F1 score across all classes. Therefore, F1-Macro retains both the biases of F1 in the single class case and introduces a further heuristic in  weighting all classes equally regardless of class prevalence.

An alternative to the F-Measure, the Receiver Operating Characteristic (ROC) curve visually presents the trade-off between Recall and Precision as a function of the decision threshold. The Area Underneath the ROC Curve (AUC) is a metric which integrates the ROC curve to return a scalar value. As \citet{Hand2009} has shown, AUC is effectively applying a cost function dependent on the False Positive Rate of the specific classifier, so systems cannot be compared if they have different False Positive Rates.

In this paper we perform an extensive empirical analysis of various classification metrics in synthetic and real settings. We advocate for using  \textbf{Informedness}, an unbiased and cognitively plausible  multi-class classification metric~\citep{Powers2003,Powers2010} for comparing classification performance of different models instead of common metrics such as accuracy and F1.  This metric avoids crediting modes exhibiting guessing or bias which distort the comparability of mainstream classification models. Informedness reports the proportion of the time a classifier makes an informed decision; that is, a decision better than bias exploitation strategies. Finally, it allows comparison between tasks of different bias or complexity, and negates the need for dataset re-balancing to `fit the metric'.

 Our main contributions are as follows:
\begin{itemize}
     \item A definition of Informedness as a classification metric suited to NLP applications
     \item Synthetic and real task comparisons of Informedness against an extensive list of classification metrics
     \item An in-depth analysis on how the use of different metrics can affect model ranking and within task understanding of model capabilities
     \item Python implementation of Informedness and Normalised Information Transfer to encourage further study within the community     
 \end{itemize}

\section{Classification Evaluation Metrics}\label{sec:rel}
We begin by defining various classification metrics and discussing their strengths and limitations. 

Metrics operate over a set of classifications, where a true class $y$ and a predicted class $\hat{y}$ are the two elements in each classification. Both $y$ and $\hat{y}$ are indications of a class from out of a set of classes $C$. The full classification output
\[
\left\{ 
  \begin{array}{c}
    \{y=C_0, \hat{y}=C_0\}, \\
    \{y=C_1, \hat{y}=C_0\}, \\
    \{y=C_1, \hat{y}=C_1\}, \\
    \vdots{}
  \end{array}
\right\}
\]
is unwieldy, so a metric is used to reduce the set more compact form, typically a single scalar value. First, the set of classifications may be considered as a Confusion Matrix (or contingency table), which is an $N\times{}N$ matrix with the columns by convention indicating the true class and the rows indicating the predicted class. Cells are assigned the number of classification events for the given actual and predicted class. In most NLP cases, creating a classification matrix is a non-destructive operation as the only information lost is the order of the classifications.

As part of our definitions, we introduce the per-class contingency table:
\begin{table}[h!]
\centering
\resizebox{\columnwidth}{!}{%
\begin{tabular}{l|lcc}
 & Class of Interest \textsubscript{c} & Other Class & \textbf{Real Class} \\ \hline
Class of Interest \textsubscript{c} & TP\textsubscript{c} & FP\textsubscript{c} & \\
Other Class & FN\textsubscript{c} & TN\textsubscript{c} & \\
\textbf{Predicted Class} & & & \\
\end{tabular}%
}
\caption{Classification Contingency Table}
\label{tab:class-cont}
\end{table}

We define this table for a class of interest \textsubscript{c}. In the binary case, this would be one of two classes and hence two tables could be created, each the 180\textdegree{} rotation of the other. In the multi-class case, there will be $c$ such matrices. 

From this table we also introduce Class Prevalence: the proportion of all samples which have a given real class, and Class Bias: the proportion of all samples which have a given predicted class. Prevalence is (TP+FN)/(TP+FN+TN+FN). Prediction Bias is (TP+FP)/(TP+FN+TN+FN).

Since an $N\times{}N$ is considered too complex to compare models, a further simplification is often used to produce a single scalar value. As this reduction is an information-destructive operation \citep{Chicco2021}, the heuristic rule \citep{Albacete2013} which the metric applies to obtain a single value will determine what that metric considers be a `good' model.

\paragraph{Accuracy:} It is defined as the proportion of correctly identified samples out of a total set of evaluation samples. Accuracy encodes the heuristic that the best model will have the most correctly predicted instances. This prior allows for the `accuracy paradox' where an uninformed model may guess the most common class artificially overestimating the generalizability score.

\begin{align}\label{accuracy} \textrm{Accuracy} = \frac{1}{S} \sum_{c=0}^{C} TP_c
\end{align}
where $C$ is the number of classes, $TP_c$ is the number True Positives for class $c$ and $S$ is the total number of samples. 

\paragraph{Balanced Accuracy:} This is a variant designed to counteract the class-frequency-weighted nature of accuracy \citep{Brodersen2010}. As shown by \citet{Chicco2021}, the binary case is equivalent to a re-scaled Informedness (see below).

\paragraph{F-Measure:}
This metric is defined as the geometric mean of the Precision and Recall of a binary classifier. 
{\small
\begin{align}\label{f1-macro}
  \textrm{F1-Macro} = \frac{1}{C} \sum_{c=0}^{C} \frac{TP_c}{TP_c+ \displaystyle \frac{1}{2} (FP_c+FN_c)}
\end{align}
}
where $TP_c, FP_c, FN_c$ denote True Positives, False Positives and False Negatives for each class $c$.  In the multi-class case (3+ classes), those are computed for each class in turn.  F1-Macro encodes the heuristic that the average of F1-Measure for all classes is a good representation of model performance. However, this has no intuitive interpretation. Additionally, as the number of negative samples increases, the number of samples which are misclassified as positive will also increase. As F1 is independent of the total number of samples, it ignores this important component of model assessment. F-Measure may be generalised to multi-class classification through micro or macro averaging. Micro-averaging sums the True Positives, False Positives, and False Negatives when calculating Precision and Recall, and is equivalent to accuracy in the uni-label case. Macro-averaging takes the arithmetic mean over Precision and Recall for every class.

\paragraph{Kappa:} This is a family of metrics which calculate the inter-annotator reliability between annotators, rather than the performance of a classifier on a task. However, they account for the probability of chance agreement. Given annotators $a_0, a_1$ they take the general form:

\begin{equation}
    k = \frac{\text{Accuracy}(a_0, a_1)-\text{Chance Agreement}(a_0, a_1)}{1-\text{Chance Agreement}(a_0, a_1)}
\end{equation}

Kappa metrics differ in how they estimate from how the chance agreement is calculated \citep{Cohen1960}. It is possible to use Kappa as a metric for classification systems by defining the system and the true labels as annotators \citep{David2007}. However, \citet{Powers2012} has shown that Kappa is unfair to models in cases where the rates of true classes and predicted classes are unequal.  

\paragraph{Informedness:} This metric treats classification evaluation as an `odds game', where a model with no predictive capability is unable to gain any credit through either \textbf{label bias} or \textbf{baseline credit}. It was first proposed in the binary case as Youden's J-statistic \citep{Youden1950} and was generalised to the multi-class case in \citet{Powers2003}. Informedness is defined as the proportion of samples for which the model guesses better than random chance. The expected value of a model which is always correct is 1, and the expected value of a model which predicts correctly x\% of the time, and guesses from the prevalence 100-x\% of the time is x. 

For a class with an empirical probability (prevalence) of $p(y=c)$, the gain (or loss) $i$ for a single prediction is computed as:

\begin{equation}\label{inf-odds}
    i(y,\hat{y}) = \begin{cases}
      \displaystyle \frac{1}{p(y=c)} \mathrm{\hspace*{1.1cm}~if~} \hat{y}=c\\[15px]
      \displaystyle -\frac{1}{1-p(y=c)} \mathrm{~~~if~} \hat{y} \neq{} c
    \end{cases}
\end{equation}
where $p(y=c)$ is the empirical probability of class $c$, calculated from the test set. 
Scores are aggregated across the whole classification set as:
\begin{equation}\label{inf-full}
    \textrm{I}=\sum_{c=0}^{C} \frac{p(\hat{y}=c)}{N}  \sum_y \mathbbm{1} 
(y=c) i(y,\hat{y})
\end{equation}

Where $\mathbbm{1}(y=c) $ is an indicator function which takes 1 when $y=c$ and 0 otherwise.

\paragraph{Mathew's Correlation Coefficient (MCC):} MCC is a measure of the correlation of the predicted classes $\hat{y}$ with the true classes $y$. Whilst its definition ensures that random guessing will score 0, for any model better than random guessing, it will not report the possibility of random chance.  MCC is dependent on the relative frequencies of classes in the test set, which makes comparison between models evaluated on different datasets impossible  \citep{Chicco2021}. Formally, MCC is defined as:
\begin{equation}
    \textrm{MCC}=\frac{\textrm{Cov}(\hat{y}, y)}{\sigma_{\hat{y}}\cdot{}\sigma_{y}}
\end{equation}

\paragraph{Normalized Information Transfer (NIT):} This information-theoretic measure reports the degree to which the classifier reduces the uncertainty of the input distribution by considering the information transfer through the classifier. It was introduced by \citet{Albacete2013}. Formally, NIT is defined as:
\begin{equation}
    \textrm{NIT} = 2^{\textrm{MI}_{\hat{y}, y} - H_{U_{y}}}
\end{equation}

Where $\textrm{MI}_{\hat{y}, y}$ is the Mutual Information of the Real and Predicted Classes, whilst $H_{U_{y}}$ is the Entropy of the Real Classes if they come from a uniform distribution.

As with Informedness, NIT considers prevalence, forcing classifiers to add Shannon Information, that is, to correctly classify samples, in order to increase the metric score.

\begin{figure*}[!t]
     \centering
     \begin{subfigure}[b]{\textwidth}
         \centering
         \includegraphics[width=\textwidth]{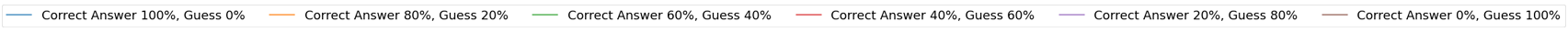}

     \end{subfigure}
     \begin{subfigure}[b]{\textwidth}
         \centering
         \includegraphics[width=\textwidth]{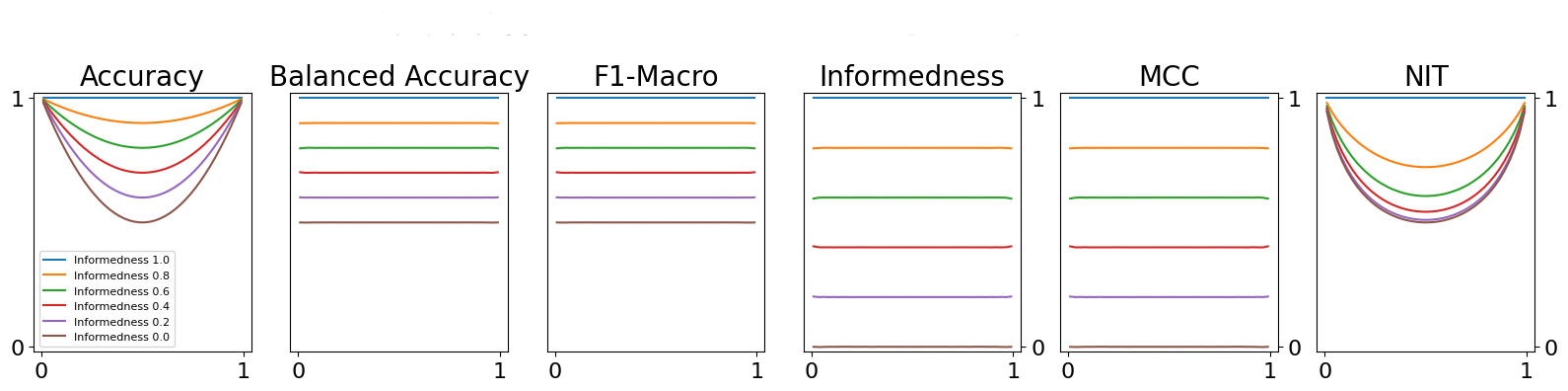}
         \caption{Binary classification case}
         \label{fig:synth_metric_binary}
     \end{subfigure}\\
     \begin{subfigure}[b]{\textwidth}
         \centering
         \includegraphics[width=\textwidth]{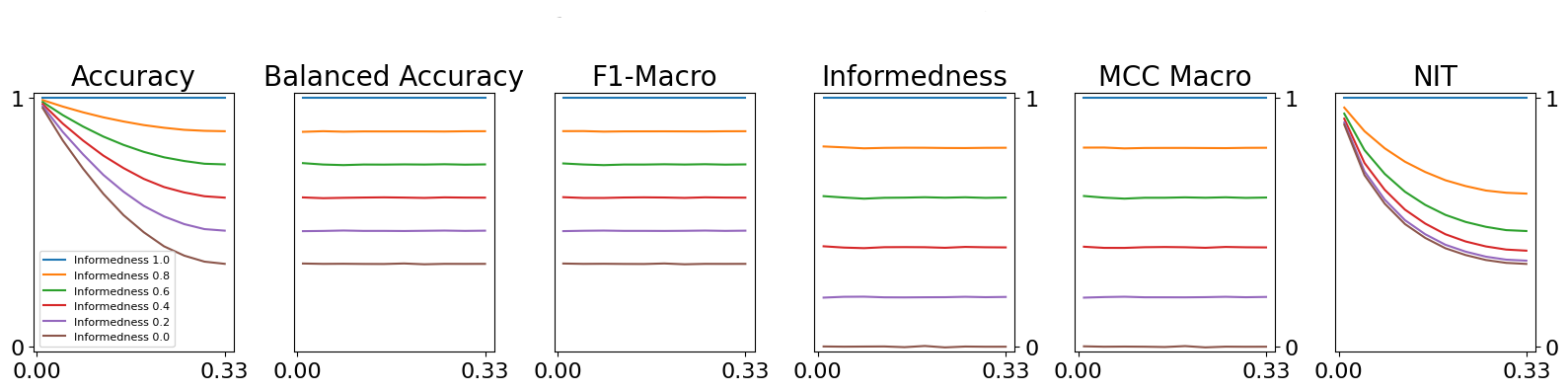}
         \caption{Multi-class classification case}
         \label{fig:synth_metric_multi}
     \end{subfigure}
     
        \caption{Accuracy, Balanced Accuracy, F1-Macro, Informedness, MCC, and NIT of the same binary (top) or multi-class (bottom) classifier as a function of the class distribution and the model's prediction capability from 0\% (Random Guess) to 100\% (Perfect). }
        \label{fig:synth_metric}
\end{figure*}

\section{Experiment 1: Metric Evaluation on a Toy Setting}\label{sec:toy}

We first compare the metrics outlined in Section \ref{sec:rel} on a toy setting, aiming to unveil the main differences between them.  We assume a simulated model as follows:
\begin{itemize}
    \item First, we sample from a uniform distribution [0,1] and then pick the correct label if the sample is smaller than model predictive power;
    \item Otherwise,  we randomly sample from the class-prevalence weighted output distribution.
    \item We score a simulated model with a fixed probability of making a correct classification
\end{itemize}
We believe this is an acceptable representation of how a reasonably designed and trained neural network would behave.

Figure~\ref{fig:synth_metric} shows the performance of a binary (top) and multi-class (bottom) classifier as a function of the class distribution and the model's predictive capacity from random guess to perfect.

In the binary case, we first observe that Accuracy becomes more distored as as the prevalence of either class increases.  On the other hand, Balanced Accuracy and F1-Macro score are robust against prevalence, but are susceptible to random chance exploitation. Surprisingly, the NIT is superficially similar to accuracy. This can be explained by the fact that when one class is far more probable than the others, the Mutual Information between a random distribution sampled from the same prior is high.

In both binary and multi-class cases MCC-Macro appears to behave exactly as Informedness. This only holds in the case where the \textit{classification ability of the model is constant across classes} \citep{Chicco2021}. We simulate model ability as a function of prevalence, so our figures do not capture this dynamic of the MCC-Macro. However, we do show that in this case Informedness correctly identifies the underlying probability of the model making an informed decision.

\section{Experiment 2: Metric Evaluation on Natural Language Understanding Tasks}

\begin{table*}[!t]
\resizebox{\textwidth}{!}{
\centering
\begin{tabular}{lcc|ccc|ccccc|c}
\toprule
\textbf{} &
  \multicolumn{2}{c|}{\textbf{Single Sentence}} &
  \multicolumn{3}{c|}{\textbf{Similarity and Paraphrase}} &
  \multicolumn{5}{c|}{\textbf{Natural Language Inference}} &
  \multicolumn{1}{l}{\textbf{}} \\ 
\textbf{Model (Metric)} &
  \multicolumn{1}{c}{\textbf{CoLA}} &
  \multicolumn{1}{c|}{\textbf{SST-2}} &
  \multicolumn{1}{c}{\textbf{MRPC}} &
  \multicolumn{1}{c}{\textbf{QQP}} &
  \multicolumn{1}{c|}{\textbf{STS-B}} &
  \multicolumn{1}{c}{\textbf{MNLI-M}} &
  \multicolumn{1}{c}{\textbf{MNLI-MM}} &
  \multicolumn{1}{c}{\textbf{QNLI}} &
  \multicolumn{1}{c}{\textbf{RTE}} &
  \multicolumn{1}{c|}{\textbf{WNLI}} &
  \multicolumn{1}{c}{\textbf{All}} \\ \midrule
DistillBERT (Acc.)                 & 79.7  & 90.5  & 84.2  & 77.4 & 51.8 & 81.4  & 81.6 & 88.6  & 57.6  & 56.3  & 74.9  \\
DistillBERT (Inform.)             & 57.0  & 81.0  & 69.4  & 77.4 & 41.6 & 72.1  & 72.5 & 77.2  & 14.7  & -43.1 & 52.0  \\
Random Guess (Acc.)           & 58.1  & 51.4  & 56.7  & 53.5 & 18.3 & 33.5  & 33.6 & 50.0  & 49.9  & 51.8  & 45.7  \\
Random Guess (Inform.)      & 01.2  & 02.8  & -01.1 & 00.0 & 01.0 & 00.1  & 00.5 & 00.0  & -00.3 & 02.0  & 00.6  \\
\midrule
$\Delta$ Accuracy         & 21.6 & 39.1	 & 27.5	 & 23.9	& 33.5 & 47.9  & 48.0 & 38.6  & 07.7  &	04.5  & 29.2 \\
$\Delta$ Informedness     & 55.8 & 78.2	 & 70.5	 & 77.4	& 40.6 & 72.0  & 72.0 & 77.2  & 15.0  & -45.1 &	51.4 \\
\bottomrule
\end{tabular}
}
\caption{GLUE Results. See \citet{wang-etal-2018-glue} for tasks details and evaluation metrics. All values are scaled by 100. `All' is a uniform weighted mean of the individual metric scores as in \url{https://gluebenchmark.com/leaderboard}.}
\label{tab:glue-metrics}
\end{table*}

Next, we compare metrics across a range of NLU tasks and show that the metric choice affects the model ranking.  First, we test on the \textbf{GLUE} Multi-Task Natural Language Understanding Benchmark. GLUE is a suite of nine NLP tasks representing a range of domains, biases, and difficulties \citep{wang-etal-2018-glue}.  Interestingly the GLUE employs  different metrics across tasks, i.e.  Accuracy, MCC, Pearson Correlation and Spearman's Correlation. MCC is a discretised version of the Pearson correlation and Spearman's Correlation is the Pearson Correlation calculated on the Rank transformation of the values. To make the continuous $ [0,5] $ STS-B task values tractable for classification metrics, we discretize into  $[0,5] \cap \mathbb{Z}$ by rounding to the nearest integer. 

We experiment with following two approaches:
\begin{itemize}
    \item {\bf Random Guess:} A `most likely' guesser, which chooses the most common class from training;
    \item {\bf DistilBERT:} We also finetune DistilBERT~\citep{Sanh2019}  for five epochs on each sub-task.
   \end{itemize}

Table~\ref{tab:glue-metrics} shows model performance across models, metrics and tasks.
For the sake of clarity, the last two lines show the difference between DistilBERT and Random Guess scores. The `All' column is a uniform-weighted mean of the metric scores across the GLUE tasks. In the case of informedness, it represents the average probability of an informed decision across all nine tasks. The use of Informedness across the GLUE tasks allows for direct comparison with the knowledge that bias is discounted. 

First, we note that sampling classes according to their prior probability (see \textit{Guess} rows) produces high accuracy scores for many tasks whilst Informedness remains very close to 0. 
This fact makes it clear that Informedness provides a more interpretable metric when it comes to evaluating model capability. 
For all tasks, we observe a lower Informedness than Accuracy. This is expected due to the properties of the metrics shown in Figure~\ref{fig:synth_metric}. 
For unbalanced tasks (CoLA, MRPC, WNLI), the gap between accuracy and Informedness is increased as Informedness removes the \textit{label bias} gain. In the three-class tasks (MNLI-M and MNLI-MM), the delta between accuracy and Informedness is reduced but still pronounced. 

WNLI is the most interesting result. DistilBERT accuracy (56.3) is a small amount (4.5) larger than random guessing which suggests a weakly predictive model. However, Informedness is strongly negative (-43.1), which suggests that the model is \textit{underperforming} the prior class distribution to a large degree. We hypothesise this is because the WNLI task is \textbf{adversarial}. We quote the GLUE authors: `Due to a data quirk, the development set is adversarial: hypotheses are sometimes shared between training and development examples, so if a model memorizes the training examples, they will predict the wrong label on corresponding development set example.' \citep{wang-etal-2018-glue} Here accuracy suggests a weak model, whilst Informedness reports the real behaviour.

Another advantage of Informedness is the possibility of direct comparison between tasks with varying bias (e.g.  CoLA and SST-2) and varying classes (e.g. CoLA and MNLI) without the need to correct for prevalence. Because MCC gives each class equal weight, it cannot be used to compare across tasks with varying class distributions \citep{Chicco2021}. Informedness and NIT support comparison between tasks, but NIT may be confusing for task comparison as it awards credit for guessing.

\section{Experiment 3: Metric Evaluation in Visual Question Answering}
Visual Question Answering (VQA) is the task of answering a question about an image and is often cast as a classification task which requires selecting a correct answer from a large set of candidate classes \citep{Antol2015}. Due to the real-world imbalances (for instance, more tables are made of wood than marble), VQA datasets have high tendencies to inherent biases, making accuracy a poor metric to use. 

In this work, we consider two VQA datasets: (1) GQA \citep{Hudson2019} and (2) KVQA \citep{Shah2019}

\begin{figure*}[!t]
     \centering
     \begin{subfigure}[b]{0.49\textwidth}
        \centering
         \includegraphics[width=0.8\columnwidth , height=11cm]{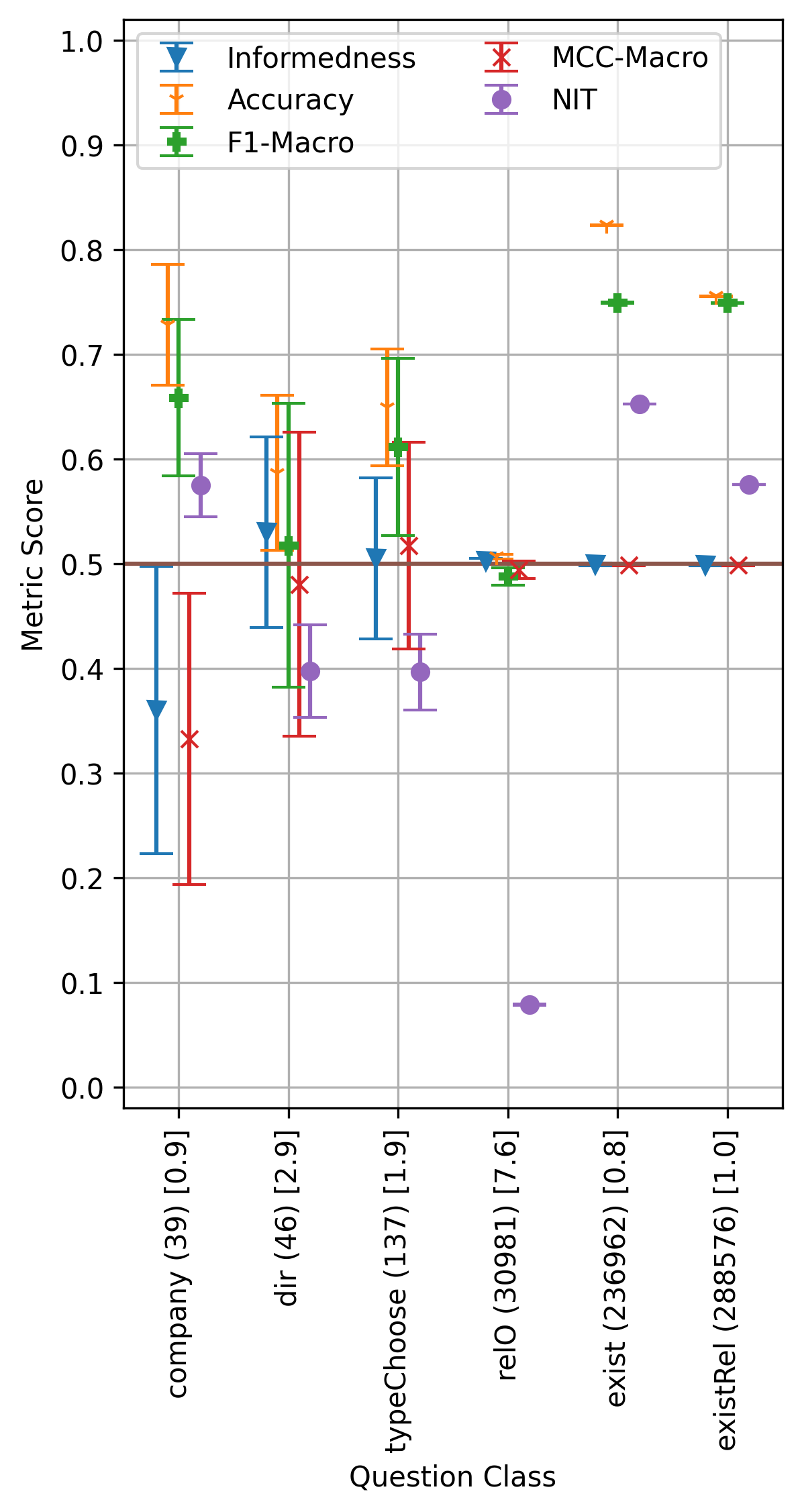}
         \caption{Unbalanced dataset \label{fig:gqa_unbalanced}}
         
     \end{subfigure}
     \begin{subfigure}[b]{0.49\textwidth}
         \centering
         \includegraphics[width=0.8\columnwidth , height=11cm]{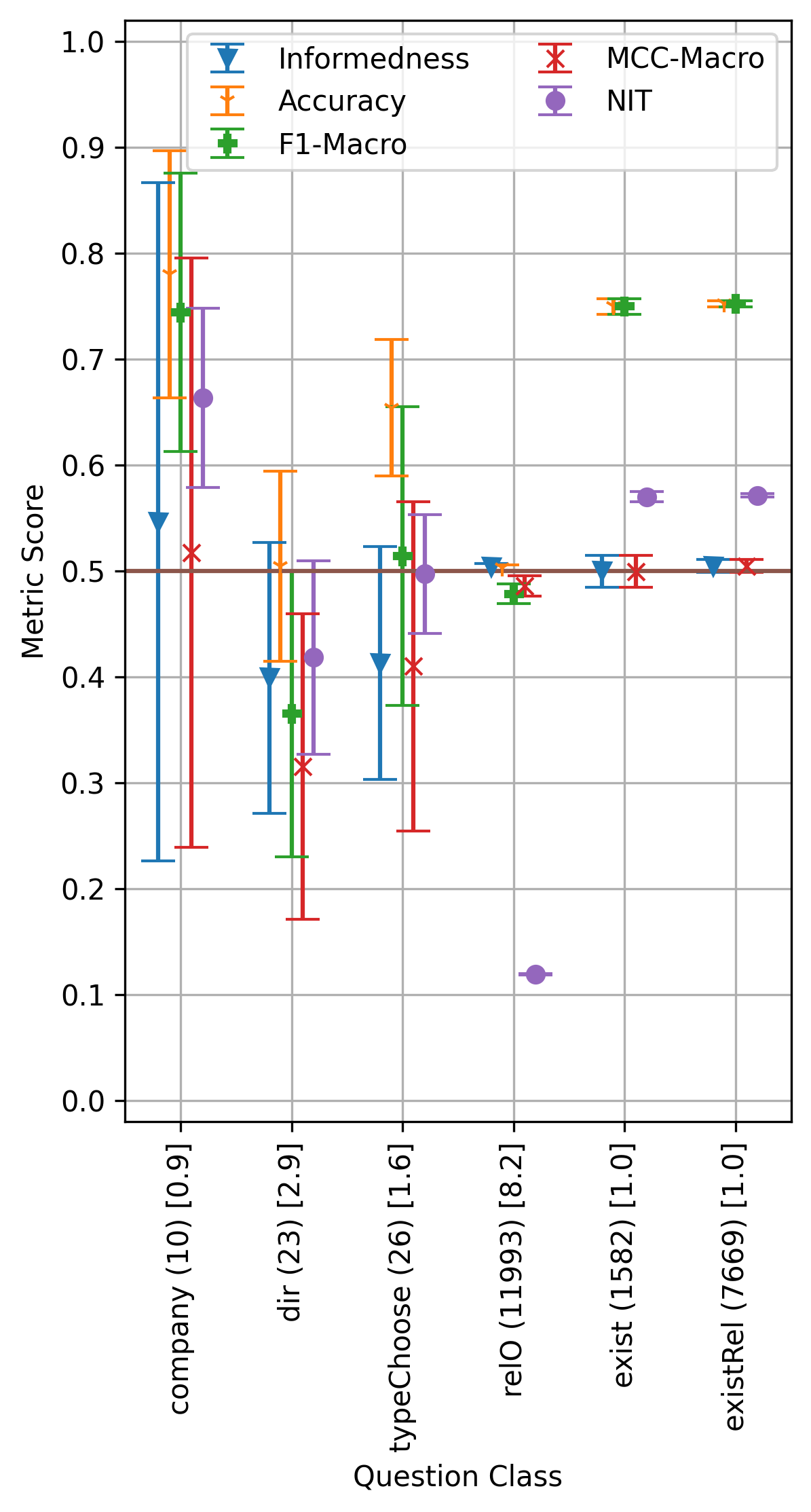}
         \caption{Balanced dataset \label{fig:gqa_balanced}}
         
     \end{subfigure}
        \caption{Metrics on GQA Unbalanced (left) and Balanced (right) validation splits. Error-bars show the standard deviation across five runs. Numbers after the question category are (question count) and [answer class entropy].}
        \label{fig:gqa_results}
\end{figure*}

\subsection{GQA}\label{sec:gqa}
We select GQA for the high variance in class count and prevalence across question types. It provides `unbalanced' and `balanced' versions. `Unbalanced' is the default dataset and features a strong prevalence skew due to real world biases towards certain classes. `Balanced' is a resampled version of dataset where the class distributions have been resampled to reduce the class imbalance.  

With GQA, we perform an intra-dataset comparison. Such a comparison is a common step in model and dataset analysis when researchers wish to compare the relative capabilities of a model on different sub-tasks. We provide a model with a predictable behaviour by simulating a 50\% probability of choosing the correct answer and a 50\% probability of sampling from the class prevalence within a question type.  For clarity, we only examine the low-frequency categories `company', `dir' and `typeChoose' and the high-frequency categories `relO', `exist', and `existRel'.  Results for a representative sub-set of the question types are shown in Figure~\ref{fig:gqa_results}.  Refer to Appendix~\ref{sec:appendix-gqa} for the full dataset results.

\begin{table*}[t!]
\centering
\resizebox{\linewidth}{!}{
\begin{tabular}{@{}l|ll|lllll@{}}
\toprule
& \multicolumn{2}{c}{\textbf{Dataset}}                         & \multicolumn{5}{c}{\textbf{Metric}}                                                     \\
\textbf{Question type} & \textbf{Classes} & \textbf{Entropy} & \textbf{Accuracy} & \textbf{F1-Macro} & \textbf{Informedness} & \textbf{MCC-Macro} & \textbf{NIT} \\ \midrule
1-Hop        & 5336 & 7.4  & 66.9 & 10.8 & 64.6 & 10.8 & 25.8 \\
1-Hop Count. & 5    & 1.1  & 79.3 & 38.9 & 58.1 & 31.5 & 58.1 \\
1-Hop Subtr. & 66   & 4.1  & 26.5 & 03.0 & 18.8 & 02.9 & 17.3 \\
Boolean      & 2    & 1.0  & 94.9 & 63.2 & 89.7 & 89.7 & 81.9 \\
Comparison   & 11   & 2.1  & 91.1 & 37.0 & 90.2 & 47.3 & 84.9 \\
Counting     & 9    & 2.1  & 80.9 & 56.1 & 75.4 & 56.2 & 61.2 \\
Intersect.   & 2    & 1.0  & 79.5 & 78.5 & 56.3 & 59.5 & 62.1 \\
Multi-Ent.   & 81   & 3.2  & 78.0 & 10.8 & 76.1 & 12.0 & 56.5 \\
Multi-Hop    & 119  & 3.6  & 87.9 & 34.8 & 87.0 & 43.9 & 68.9 \\
Multi-Relat. & 4104 & 6.8  & 75.4 & 11.7 & 73.7 & 12.1 & 38.1 \\
Spatial      & 1260 & 10.0 & 19.9 & 07.4 & 18.6 & 09.2 & 16.3 \\
Subtract.    & 93   & 5.9  & 39.8 & 36.6 & 45.9 & 34.3 & 08.6 \\ \bottomrule
\end{tabular}
}
\caption{Model performance on KVQA across metrics.}
\label{tab:kvqa}
\end{table*}

First, we have many cases where Accuracy, Balanced Accuracy and F1-Macro are 75\% on binary questions. This baseline credit makes it hard to compare between model performance, which is calibrated to be uniform, across dataset sub-tasks. Practically, we are not able to use Accuracy, F1-Macro, or NIT to look at `typeChoose' questions and see if the model is as strong as on `existRel'. Meanwhile, MCC-Macro and Informedness converge on the correct value (0.5) even with the 46 samples in `dir' question type. The `dir' case demonstrates how the deletion of samples to create a more uniform prevalence is not required with sophisticated metrics. That is, Informedness and MCC are closer to the true value for `dir' with the unbalanced sample than with the balanced one. Meanwhile, the balanced dataset has only a minor effect on accuracy and F1-score, with `dir' and `typeChoose' questions being slightly closer to an unbiased score. This reinforces our hypothesis that dataset balancing is not the correct approach to evaluation.

For the questions with many samples (`relO', `exist', and `existRel'), all metrics have low variance. For `exist', and `existRel', F1-Macro and Accuracy converge on 0.75, which reflects correctly predicting a binary task half the time, and randomly guessing the other half. For the `relO' question class, Accuracy and F1-Macro tend to the true proportion of the time the model is predicting the correct answer, but this can be attributed to the higher entropy for this class of questions. The same behaviour can be observed for additional question types in \autoref{sec:appendix-gqa}. 

These experiments show that that Informedness automatically accounts for prevalence imbalance and provides a better assessment of the model capability. Whilst MCC appears similar, it over-punishes classifiers which have variable per-class performance \cite{Chicco2021}, which we do not believe is in line with desired characterises of classifiers in NLP.

\subsection{KVQA}
Having established metric characteristics through controlling model performance, we now move to model evaluation in the wild. First, the KVQA dataset \citep{Shah2019} provides multiple question type attributes for each question. The task requires reasoning over retrieved knowledge graph facts as well as arithmetical operations. For modelling,  we select `REUNITER', a simple yet effective transformer based model  \citep{Vickers2021}, and re-evaluate it with informedness. 

We are interested in this case for the opportunity to have a metric which allows comparison \textit{within} a dataset \textit{between} subsections with different class distributions. We present results across unbiased metrics  Informedness, MCC-Macro and NIT \citep{Powers2003, Chicco2021, Albacete2014} along with accuracy grouped by question type in Table~\ref{tab:kvqa}. 

The `1-Hop' category is a superset of many question types requiring a single KG fact to answer. This question type is scored very differently across all metrics but the difference between Informedness (64.6\%), MCC-Macro (10.8\%) and NIT (25.8\%) is especially striking given the agreement between Informedness and NIT in the synthetic case from Section~\ref{sec:gqa}. This range indicates the model is doing well in general: if it were guessing from a prior, it would have an Informedness of zero. The difference can be explained by the different dynamics of Informedness and MCC raised above. The model is much better than random chance at predicting certain popular classes, but struggles with low-frequency obscure classes. This is supported by a high accuracy at the same time as a low F1-Macro (12.9). In this case, F1-Macro, MCC, and NIT harshly and unfairly penalize the model.

Looking at the `Intersection' type, we see the opposite behaviour. Accuracy and F1-Macro are all fairly high (78.5 and above) while Informedness is rather low (56.3). This means that Accuracy and F1-Macro exaggerate the predictive power of the model for this type of question. The similar score of MCC-Macro (59.5\%) to Informedness indicates that the model has even performance across classes. 

Interestingly, accuracy reports that the model is poor at `subtraction' questions, which Informedness is much higher (45.9).
We hypothesise this is because (1) transformer models are not good at arithmetic without extensive task-specific pretraining and (2) the high number of output labels will have lower \textit{baseline credit}. 

Through the use of Informedness, we come to a different conclusion of the relative strengths of the model. We find that the model has better mathematical ability than accuracy indicated, whilst the ability to reason over intersectional facts is much poorer than accuracy reports. For example, this could lead to focus on improving this sub-task in the future.

Meanwhile, we have the issue that both Informedness and NIT are proposed as suitable metrics for reporting the cross-task capability of different classifiers, but they report divergent scores and sub-task rankings. This is because both metrics target different criteria: NIT the transmission of information from the true labels to the predicted labels, and Informedness the probability of an informed decision. We propose that Informedness is a more intuitive measure for NLP, and refer to Section \ref{sec:toy} for a toy example demonstration.

\section{Experiment 4: Metric Evaluation on Formality Control for Spoken Language Translation}

\begin{table}[!t]
\centering
\resizebox{\linewidth}{!}{
\begin{tabular}{rcc}
\toprule
& Off-the-shelf MT & Formality-aware MT \\
\midrule
Accuracy          & 50.0 & 95.4 \\
Balanced Accuracy & 50.0 & 95.3 \\
F1-Macro          & 49.2 & 95.4 \\
Informedness      & 00.0 & 91.8 \\
\bottomrule
\end{tabular}
}%
\caption{Metric scores on Formality Control for Spoken Language Translation (En-De) between off-the-shelf and formality-aware MT systems.}
\label{tab:formality-control}
\end{table}

In the last set of experiments, we consider a contextual task involving machine translation (MT). 
The \textit{Special Task on Formality Control for Spoken Language Translation} \citep{anastasopoulos-etal-2022-findings} evaluates an MT model to correctly express the desired formality (either \textit{formal} or \textit{informal}) in its translation hypotheses. Focusing on the English-to-German language pair, we use the winning system proposed by \citet{vincent-etal-2022-controlling}. The model is trained to recognise a formality token to generate adequate translations, and an off-the-shelf formality-unaware MT model on the test set provided by the organisers.  We report accuracy, Balanced accuracy, F1-Macro and Informedness on the English-to-German test set.

Table~\ref{tab:formality-control} displays metric scores between off-the-shelf and formality-aware MT systems. 
We see that the model with no knowledge of the formality is still able to achieve accuracy and F1-score of around 0.5, which seems to mean that the model is able to correctly produce a translation with correct formality 50\% of the time.
Meanwhile, Informedness drops to zero. As the dataset is balanced, this is a product of Informedness removing baseline credit making it a more suitable choice as an evaluation metric.

Overall, Informedness provides a better and more interpretable measure of the system capability to model the task. This demonstrates that Informedness can be used as an effective tool for comparing two different systems.

\section{Discussion}

\subsection{Limitations of current metrics} 
The results obtained across all experiments highlight that widely-used metrics (e.g. Accuracy,  F1-Macro) for classification evaluation in NLP feature biases which suggest higher performance than either intuitive reasoning or information theory support.  Importantly, this bias makes comparing classifiers across tasks with different class distributions impossible. 

Additionally, through the analysis of a real model on the KVQA task, we showed that traditional metrics are not suited to intra-dataset analysis when evaluating a single model's performance across various sub-tasks. This is highly problematic, as knowing if a model is better at a particular sub-task such as the sub-tasks of addition or syntactic parsing is crucial for model analysis. 

\subsection{Improving Evaluation of Classification Tasks in NLP}

Across all experiments, we found that Informedness better captures model generalizability than all other metrics. Given this finding and the main limitation of popular metrics such as Accuracy and F1 across different NLP tasks,  \textit{we encourage the community and practitioners to consider reporting Informedness alongside metrics such as Accuracy and F1 in future experiments and analyses}.\footnote{For a discussion of the limitations of Informedness,  see Limitations section.}

\section{Conclusion}

We have presented an extensive empirical analysis of various classification metrics across a wide range of tasks including NLU, VQA and MT with controlled formality.  Our experiments demonstrated that the use of a class-invariant metric, Informedness, allows for a fairer ranking and understanding of model generalization capacity. 

Whilst we find that Informedness is the most intuitive metric, we also found that it is also the fairest in driving inter and intra-model comparisons. 

Finally, we provide \texttt{sklearn.metrics} style implementations of both NIT and Informedness, previously unavailable in Python

We hope that our work is the first step towards rethinking the way NLP classification systems are evaluated in the future and will raise awareness to the community.

\newpage
\section*{Limitations}
Informedness cannot fully represent all of the characteristics of a classification system within a single scalar value. It assumes that the distribution of classes in the training and test set are identical. This assumption is used to determine the loss and gain for a particular class according to the distribution in the test set. However, we allow for train class distributions to be passed to our implementation of Informedness.

In this work, we further assume that an uninformed model will reproduce the training distribution. In the case that models are poorly parameterised, or the testing set is very small, this may not be the case. This could lead to models which are not using the input data to have Informedness scores other than zero. Likewise, systems which use strategies such as `guess the most common' may have Informedness scores other than zero.

Informedness is sensitive to the number of evaluation samples, which may result in less stable estimation of model's performance in situations with low numbers ($<50$) of examples. We consider that all metrics are subject to this and that it is reasonable to expect that evaluation is performed on sizeable test sets.

\section*{Acknowledgements}
This work was supported by the Centre for Doctoral Training in Speech and Language Technologies (SLT) and their Applications funded by UK Research and Innovation [grant number EP/S023062/1].

\newpage
\bibliography{anthology,custom}
\bibliographystyle{acl_natbib}

\onecolumn

\appendix

\section{GQA Full Comparison}
\label{sec:appendix-gqa}

\begin{figure}[ht]
\centering
     \includegraphics[width=\textwidth]{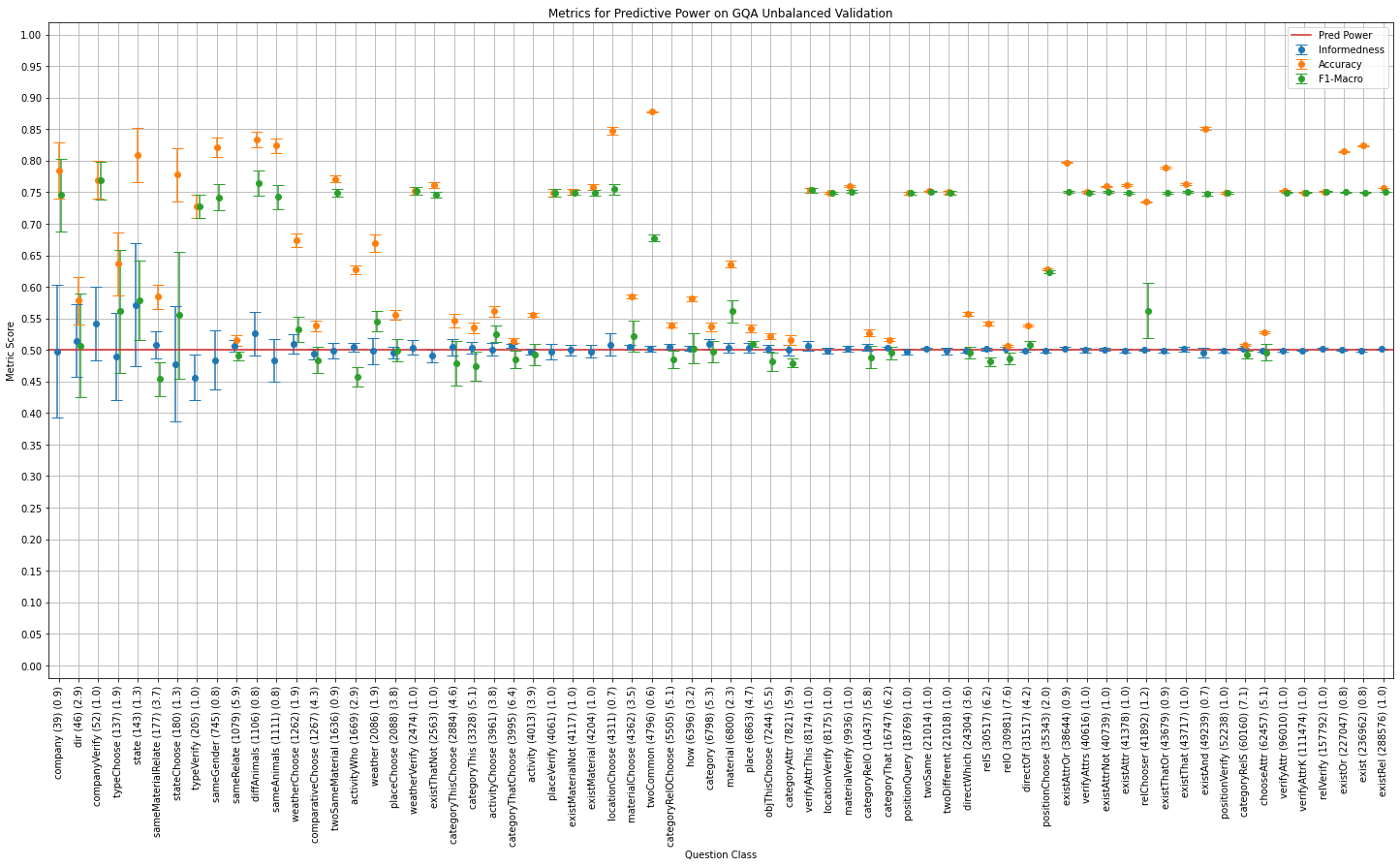}
      \caption{Metrics on GQA Unbalanced. Questions are grouped by reasoning type annotation on the X axis and sorted by count. X axis labels gives the reasoning type, the number of samples, and the entropy of the answer class distribution}
       \label{fig:gqa_unbalanced_full}
\end{figure}

\begin{figure}[ht]
\centering
     \includegraphics[width=\textwidth]{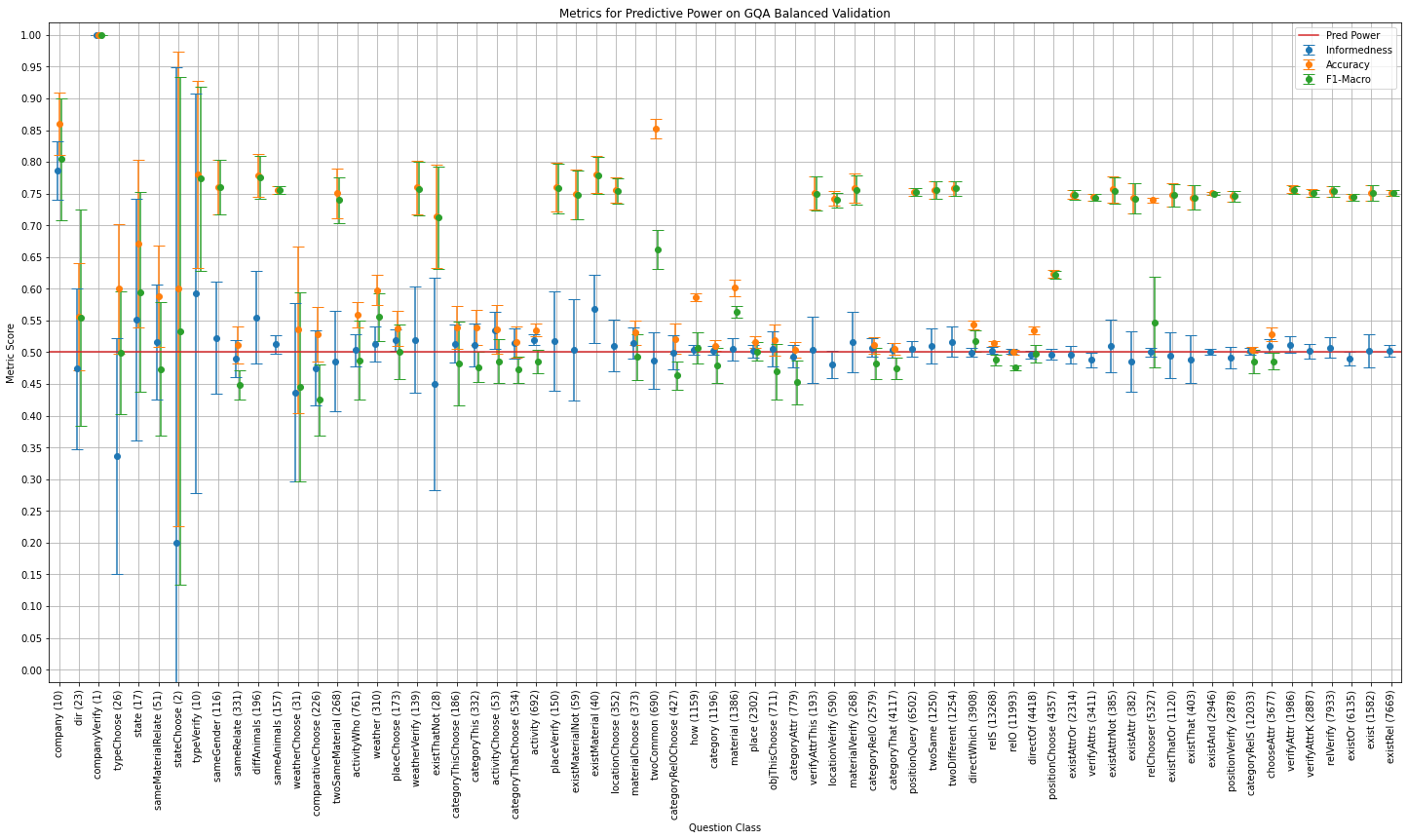}
      \caption{Metrics on GQA Balanced. Questions are grouped by reasoning type annotation on the X axis and sorted by count in GQA Unbalanced for comparison. X axis labels gives the reasoning type, the number of samples, and the entropy of the answer class distribution}
       \label{fig:gqa_balanced_full}
\end{figure}

\end{document}